\documentclass[10pt,twocolumn,letterpaper]{article}
\usepackage{multirow} 
\usepackage{float}
\usepackage{bbding}
\usepackage{cvpr}              
\usepackage{amsmath}
\usepackage[accsupp]{axessibility}
\usepackage[dvipsnames]{xcolor}

\definecolor{cvprblue}{rgb}{0.21,0.49,0.74}
\usepackage[pagebackref,breaklinks,colorlinks,citecolor=cvprblue]{hyperref}

\def\paperID{209} 
\def\confName{CVPR}
\def\confYear{2024}

\title{FlowVQTalker: High-Quality Emotional Talking Face Generation through Normalizing Flow and Quantization}

\author{
Shuai Tan, Bin Ji, and Ye Pan\thanks{Corresponding author.} \\
Shanghai Jiao Tong University \\
{\tt\small \{tanshuai0219,bin.ji,whitneypanye\}@sjtu.edu.cn}
}

\begin{document}

\maketitle
\begin{abstract}
Generating emotional talking faces is a practical yet challenging endeavor. To create a lifelike avatar, we draw upon two critical insights from a human perspective: 1) The connection between audio and the non-deterministic facial dynamics, encompassing expressions, blinks, poses, should exhibit synchronous and one-to-many mapping. 2) Vibrant expressions are often accompanied by emotion-aware high-definition (HD) textures and finely detailed teeth. However, both aspects are frequently overlooked by existing methods. To this end, 
this paper proposes using normalizing \textbf{Flow} and \textbf{V}ector-\textbf{Q}uantization modeling to produce emotional \textbf{talk}ing faces that satisfy both insights concurrently (\textbf{FlowVQTalker}). Specifically, we develop a flow-based coefficient generator that encodes the dynamics of facial emotion into a multi-emotion-class latent space represented as a mixture distribution. The generation process commences with random sampling from the modeled distribution, guided by the accompanying audio, enabling both lip-synchronization and the uncertain nonverbal facial cues generation. Furthermore, our designed vector-quantization image generator treats the creation of expressive facial images as a code query task, utilizing a learned codebook to provide rich, high-quality textures that enhance the emotional perception of the results. Extensive experiments are conducted to showcase the effectiveness of our approach.
\end{abstract}
\section{Introduction}
\label{sec:intro}
Talking face generation has garnered growing interest due to its immense potential in various contexts, including virtual reality, filmmaking, and online education~\cite{pataranutaporn2021ai}. While existing research has made significant strides in improving lip-synchronization~\cite{chen2018lip, vougioukas2020realistic, tian2019audio2face, wang2023seeing}, a notable oversight is the neglect of expressive facial expressions and diverse head poses, which are integral components of creating a lifelike and captivating avatar~\cite{tan2023emmn}. Consequently, one can readily distinguish such avatars from real humans.

\begin{figure}[t]
  \centering
  \includegraphics[width=1\linewidth]{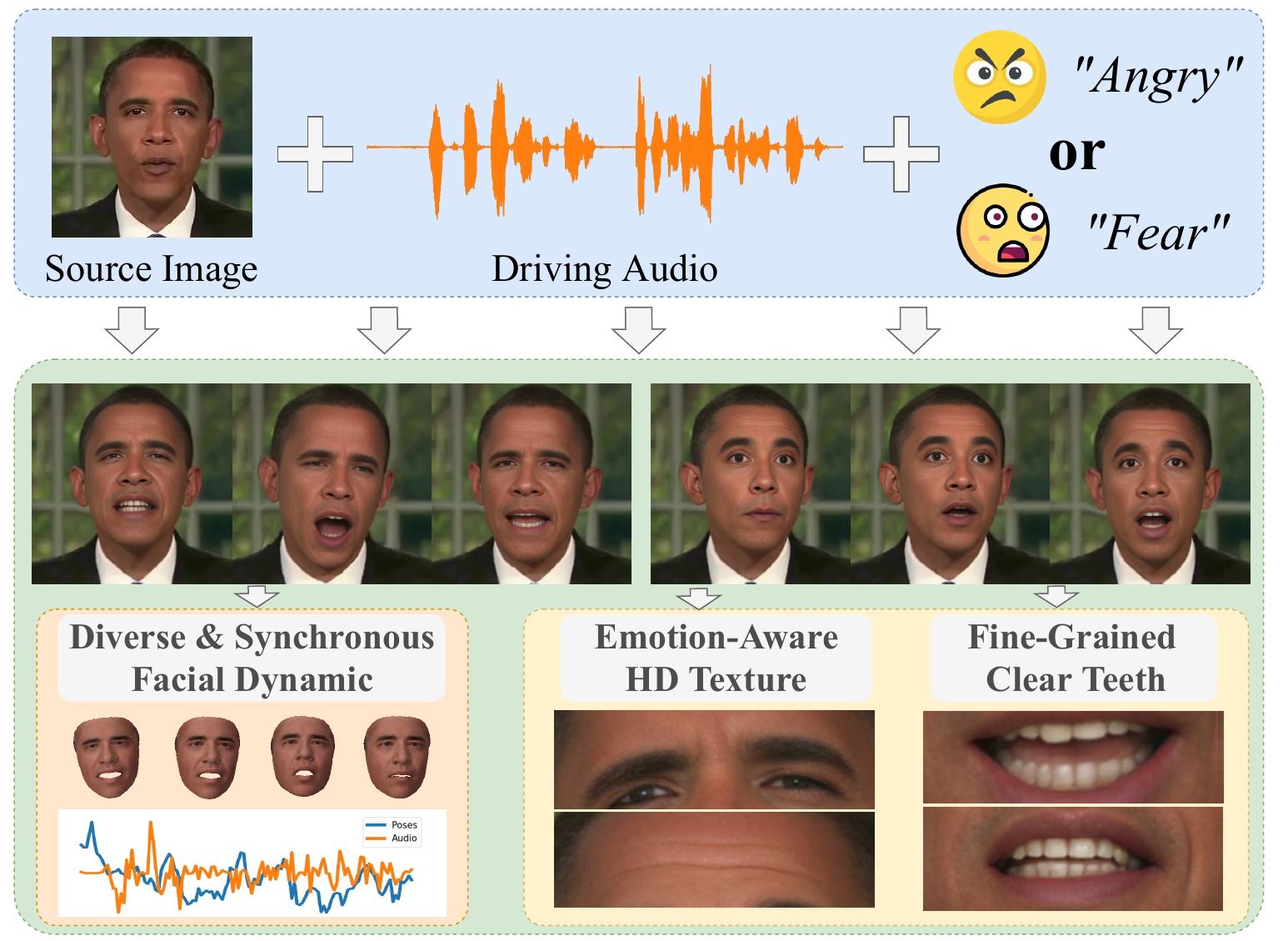}
    \caption{Example animations produced by FlowVQTalker. Given a source image and a driving audio, FlowVQTalker creates talking face video, complete with \textit{\textbf{diverse and synchronous facial dynamic}}, \textit{\textbf{emotion-aware HD texture}} and \textit{\textbf{fine-grained clear teeth}}.}
    \label{fig:teaser}
\end{figure}

To address this, we initially identify two vital observations for natural-looking talking heads from a human perspective: 1) In reality, non-verbal facial cues exhibit inherent variability, rendering them non-deterministic in nature~\cite{FaceDiffuser_Stan_MIG2023}. Therefore, the mapping from input audio to generated video constitutes a one-to-many relationship~\cite{tang2022memories}, where one audio clip can manifest in multiple plausible visual results owing to the fluid emotions, blinks, and head poses. 2) Authentic expressions not only retain the identity of source image but also feature intricate textures, such as wrinkles that intensify the expressiveness of the desired emotion. As depicted in~\cref{fig:teaser}, vertical lines between the eyebrows commonly accompany anger, while horizontal wrinkles on the forehead are associated with fear or surprise~\cite{faigin2012artist}. Furthermore, a high-quality video should encompass clear teeth. The observations succinctly encapsulate the key considerations in the avatar generation process, guiding the direction of progress and enhancement towards more engaging and realistic talking face synthesis.

Recent efforts have focused on modeling facial expressions~\cite{tan2024saas, tan2023style2talker, goyal2023emotionally} and head motions.~\cite{gan2023efficient, SanjanaSinha2022EmotionControllableGT, peng2023emotalk} rely on discrete emotional labels to prompt emotions, while~\cite{ji2022eamm, liang2022expressive, ma2023styletalk} introduce an emotion reference video to suggest the desired expression.~\cite{wang2021audio2head, wang2022one, RanYi2020AudiodrivenTF} infer head poses from audio via a time series analysis model~\cite{lim2021time}. Despite their contributions in enhancing the expressiveness of generated avatars, these methods still exhibit certain limitations: 1) Since non-verbal facial dynamics exhibit weak correlations with audio, such deterministic models tend to produce fixed and unrealistic outputs, lacking the diversity. 2)
Low-resolution image generator~\cite{ji2022eamm, tan2023emmn, ma2023styletalk} struggle to capture emotion-aware textures and clear teeth. Additionally, they face challenges in maintaining consistency between the generated video and source image~\cite{wang2023emotional, oorloff2023robust}. As a result, these methods have not addressed the two core observations mentioned earlier.

In this paper, we propose a novel method called FlowVQTalker to generate vibrant emotional talking head videos that meet: (1) diverse outputs encompassing various facial dynamics that respond to the driving audio and the emotional context. (2) preservation of the identity information from the source image, accompanied by the presentation of rich emotion-aware textures and clear teeth to enhance expressive performance and video quality. FlowVQTalker is comprised of the Flow-based Coeff. Generator (FCG) and Vector-Quantized Image Generator (VQIG), which are interconnected via the coefficients of 3D Morphable Models (3DMM)~\cite{deng2019accurate}. Within the FCG, we devise ExpFlow and PoseFlow for expression and pose coefficient modeling, built upon the generative flow model~\cite{Rezende_Mohamed_2015}. To elaborate, ExpFlow establishes an invertible transformation between emotional expression coefficients and latent codes, which are then mapped into the Student's $t$ mixture model (SMM). Each mixture component of the SMM encodes features for an emotion class, wherein latent codes within the same component represent the same emotion while differing in nonverbal facial cues. During inference, we stochastically sample latent codes from the corresponding SMM component, ensuring the \textbf{\textit{diversity}} of generated expressions. Furthermore, the one-to-one and bijective relationship between expression coefficients and latent codes enables us to achieve \textbf{\textit{emotion transfer}} by providing an emotion reference. PoseFlow employs a similar technique to ExpFlow but incorporates specific modifications for handling pose-related aspects, as detailed in~\cref{sec:3.2}.

On the other hand, VQIG approaches the synthesis of fine-grained textures and teeth from a fresh perspective. We regard image rendering as a code query task within a learned codebook. This perspective is inspired by the capabilities of the codebook in VQ-GAN~\cite{Esser_Rombach_Ommer_2021}, which excels at preserving emotion-aware texture information and supplying a wealth of visual elements for generating top-quality faces. However, the vanilla VQ-GAN framework grapples with preserving identity information when appearing frequent spatial transformations. To this end, we extract features from source image to complement the motion synthesis, facilitating \textbf{\textit{high-fidelity}} and \textbf{\textit{expressive}} talking faces creation.

In summary, our contributions are outlined as follows:
\begin{itemize}
    \item We introduce FlowVQTalker, a system capable of generating emotional talking face videos with diverse facial dynamics and fine-grained expressions.
    \item We harness the generative flow model to forecast non-deterministic and realistic coefficients. To the best of our knowledge, we are the pioneers in applying normalizing flow for emotional talking face generation.
    \item The visual codebook within our proposed VQIG enriches the textures with emotion-aware HD details, thereby enhancing expressiveness and elevating video quality.
    \item Extensive experiments demonstrate that our FlowVQTalker outperforms the competing methods in both quantitative and qualitative evaluation.
\end{itemize}

\section{Related Work}
\label{sec:related_work}

\subsection{Audio-Driven Talking Face Generation}
Existing audio-driven talking face generation methods can be broadly categorized into reconstruction-based methods and intermediate representation-based methods. On the one hand, the former~\cite{chen2018lip, Song_Zhu_Li_Wang_Qi_2019, HangZhou2019TalkingFG} typically involve mapping inputs from different modalities (e.g., audio and images) to corresponding features using encoders. Subsequently, these features are decoded to produce the talking faces. For example, Prajwal~\etal~\cite{prajwal2020lip} train their Wav2Lip following adversarial process~\cite{mirza2014conditional}, where the generator, based on an encoder-decoder architecture, reconstructs videos from the extracted features, and the lip-sync discriminator~\cite{chung2017out} is employed to improve lip-synchronization. On the other hand, the intermediate representations, like landmarks~\cite{chen2019hierarchical, zhou2020makelttalk, zakharov2019few, zhong2023identity} and dense motion fields~\cite{wang2021audio2head, wang2022one}, are leveraged to bridge the modality gap. Zhou~\etal~\cite{YangZhou2020MakeltTalkST} proposes a two-stage model to predict facial landmarks from audio, which subsequently serve as a condition for video synthesis. In contrast, we adopt 3D Morphable Model (3DMM)~\cite{deng2019accurate} as the bridge, as it offers better decoupling and control over expression, pose, and identity. Moreover, these methods neglect incorporation of expressive emotions into the generated faces, which is the main focus of our work.

\subsection{Emotional Talking Face Generation}
In the pursuit of achieving lifelike and emotionally expressive talking face generation, an increasing number of studies are incorporating emotion as a crucial element. One prevalent category of emotion sources in current methods is one-hot labels~\cite{SanjanaSinha2022EmotionControllableGT, karras2017audio, SefikEmreEskimez2020SpeechDT, Peng_Wu_Song_Xu_Zhu_Liu_He_Fan_2023, Wang_Zhao_Liu_Xu_Li_Li_2023, gan2023efficient, pan2023emotional, pan2024expressive}. Wang~\etal~\cite{wang2020mead} release an emotional audio-visual dataset MEAD and aligns neutral images with emotional states through the guidance of one-hot emotion labels using a U-Net. However, such deterministic models are prone to producing fixed expressions. Instead, our approach involves mapping different expressions into the latent distribution of a mixture model through normalizing flow~\cite{Rezende_Mohamed_2015, ji2023stylevr}. This enables us to sample a range of diverse expressions from the modeled distributions during inference. Recent works~\cite{wang2023progressive, ji2022eamm, Xu_Zhu_Zhang_Han_Chu_Tai_Wang_Xie_Liu_2023, liang2022expressive} focus on emulating the speaking styles of given reference, thereby conveying more diverse expressions. Ma~\etal~\cite{ma2023styletalk}, for instance, extract style codes from the 3DMM expression coefficients and generate lip movements synchronized with audio. Leveraging the forward process of normalizing flow, our method readily identifies the precise latent code corresponding to the given coefficients in the modeled distribution, thus achieving controllable emotional transfer.

\subsection{Vector-Quantized Codebook}
Vector-Quantized Network~\cite{Oord_Vinyals_Kavukcuoglu, Esser_Rombach_Ommer_2021} learns a codebook to store quantized features extracted from an autoencoder, significantly enhancing image modeling. Due to its effectiveness in replacing extracted features with quantized ones, this approach has demonstrated its remarkable potential in image restoration~\cite{zhou2022towards, gu2022vqfr, wang2022restoreformer} and talking face generation~\cite{Wang_Liang_Zhou_Tang_Wu_He_Hong_Liu_Ding_Liu, Xing_Xia_Zhang_Cun_Wang_Wong_2023, Wang_Zhao_Zhang_Zhang_Shen_Zhao_Zhou_Group_Group}. Ng~\etal~\cite{Ng_Joo_Hu_Li_Darrell_Kanazawa_Ginosar_2022} store facial motion in a discrete codebook and generate potential responsive motions of listeners during conversations. Wang~\etal~\cite{Wang_Liang_Zhou_Tang_Wu_He_Hong_Liu_Ding_Liu} learn position-invariant quantized local patch representations and employ a transformer to achieve face reenactment. However, both methods construct speaker-specific codebooks, which face challenges 
to generalize on arbitrary identities and facial motions. In contrast, our approach utilizes a generic codebook capable of representing a wide range of identities and facial expressions, enabling the production of high-quality talking face videos with rich emotional facial textures. To the best of our knowledge, we are the pioneers in employing normalizing flow and vector-quantized codebooks for speech-driven emotional facial animation.

\section{Method}

\begin{figure*}
  \centering
  \begin{subfigure}{0.38\linewidth}
    \includegraphics[width=1\linewidth]{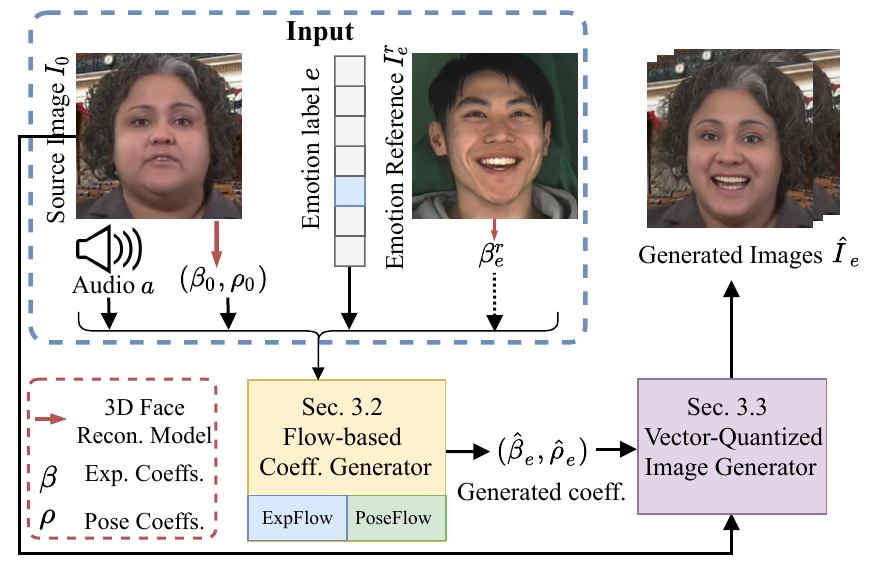}
    \caption{The pipeline of our proposed FlowVQTalker}
    \label{fig:pipeline}
  \end{subfigure}
  \hfill
  \begin{subfigure}{0.61\linewidth}
\includegraphics[width=1\linewidth]{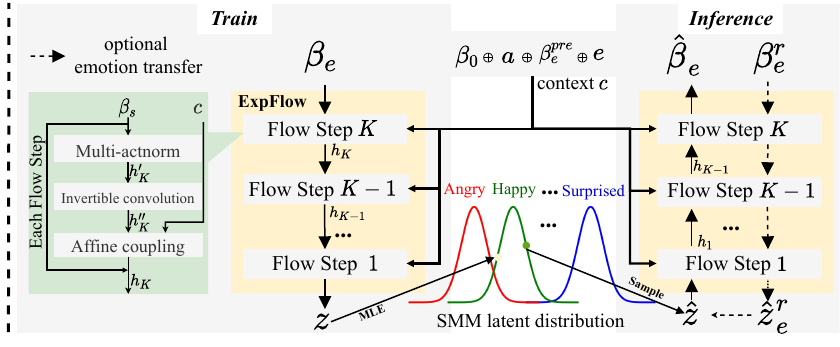}
 \caption{The framework of ExpFlow and train/inference workflow.}
    \label{fig:flow}
  \end{subfigure}
  \caption{The overview of our proposed FlowVQTalker. (a) Main pipeline. Given a source image $I_0$, audio $a$ and emotion label $e$, Flow-based Coeff. Generator (~\cref{sec:3.2}) which consists of ExpFlow and PoseFlow generates synchronized emotional coefficients $(\hat{\beta_e}, \hat{\rho_e})$. ExpFlow also supports emotion transfer by inputting an additional emotion reference $I^r_e$. Subsequently, Vector-Quantized Image Generator (\cref{sec:3.3}) takes as input the generated coefficients $(\hat{\beta_e}, \hat{\rho_e})$ and source image $I_0$, producing emotional talking face frames $\hat{I}_e$. (b) The framework of ExpFlow and train/inference workflow. ExpFlow is composed of $K$ flow steps, each containing three subsections: multi-actnorm, invertible convolution, and an affine coupling layer. Based on the property of normalizing flow~\cite{rezende2015variational}, our proposed ExpFlow is bijective. To provide a clearer understanding, we separately represent the training (left) and inference (right) workflows, but note that they run on the same network structure. During training, the ground truth emotional coefficient $\beta_e$ is passed through the forward process (left) of ExpFlow, leading to the conversion into a latent code $z$, which is then mapped into the SMM latent space using Maximum Likelihood Estimation (MLE). During inference, we sample $\hat{z}$ from the distribution corresponding to emotion $e$, which is then fed into the reverse process (right) of ExpFlow to generate $\hat{\beta}_e$. Besides, as indicated by the dotted arrows, $\hat{z}$ can also be obtained by inputting emotion reference $\beta^r_e$ into forward process of ExpFlow, achieving optional emotion transfer. Both training and inference are conditioned on contextual information, including the source coefficient $\beta_0$, driving audio $a$, previous coefficients $\beta^{pre}_e = \beta^{t-\tau:t-1}_e$ and emotion label $e$, where $t$ and $\tau$ refer to current frame index and previous frame length. Note that we employ an audio encoder and an emotion encoder to obtain corresponding features from $a$ and $e$ before the concatenation operator, and the detailed structures are in supplementary material.}
  
\end{figure*}

\subsection{Overview}
\cref{fig:pipeline} illustrates the pipeline of FlowVQTalker, which is designed to create emotional talking head videos based on a source image $I_0$, driving audio $a$ and emotion label $e$. We also achieve emotion transfer by introducing an emotion reference $I^r_e$. Basically, we leverage the expression coefficients $\beta \in \mathbb{R}^{64}$ and pose coefficients $\rho \in \mathbb{R}^{6}$ of 3D Morphable Models (3DMM)~\cite{blanz1999morphable} to represent facial dynamics. Specifically, we input $\{(\beta_0, \rho_0), a, e\}$ or $\{(\beta_0, \rho_0), a, \beta^r_e\}$ (for emotion transfer) into Flow-based Coeff. Generator (\cref{sec:3.2}), comprising ExpFlow and PoseFlow, to predict emotional facial dynamics $(\hat{\beta}_e, \hat{\rho}_e)$. By integrating the designed Vector-Quantized Image Generator (\cref{sec:3.3}), we generate emotional talking head frames $\hat{I}_e$ from the generated $(\hat{\beta}_e, \hat{\rho}_e)$ and the source image $I_0$. The following subsections will delve into the finer details of each module.

\subsection{Flow-based Coeff. Generator}
\label{sec:3.2}

\noindent \textbf{Preliminary of Generative Flow Model.} Normalizing flow~\cite{Rezende_Mohamed_2015} serves as a generative model with a primary advantage: it can effectively model a complex distribution $\mathcal{X}$ by leveraging a simple, fixed base distribution $\mathcal{Z}$ through an invertible and differentiable nonlinear transformation $f$. This reversibility of $f$ implies that $z \in \mathcal{Z}$ can be readily obtained from $x \in \mathcal{X}$ using an inverse process denoted as $f^{-1}$, constituting a \emph{normalizing flow}. When dealing with highly complex distributions, it is often necessary to employ a series of multiple flow steps $\{f_n\}^K_{n=1}$ to achieve the desired distribution modeling: $f = f_K \circ \cdots \circ f_{2} \circ f_{1}$. Formulaically, for a given complex distribution $\mathcal{X} \sim p_{\mathcal{X}}$, a simple distribution $\mathcal{Z} \sim p_{\mathcal{Z}} $ and hidden distributions $\mathcal{H}_n$, the transformations among these can be represented as:
\begin{equation}
{z} \stackrel{{f}_1}{\longleftrightarrow} {h}_1 \stackrel{{f}_2}{\longleftrightarrow} {h}_2 \cdots \stackrel{{f}_K}{\longleftrightarrow} {x}
\end{equation}
\begin{equation}
x = f(z) =  f_K(f_{K-1}(...f_1(z)))
\end{equation}
\begin{equation}
z = f^{-1}(x)= f^{-1}_1(f^{-1}_{2}(...f^{-1}_K(x)))
\end{equation}
Given $\left(\frac{\partial z_k}{\partial z_{k-1}}\right)$ as the Jacobian matrix of $f^{-1}_n$ at $x$, the log-likelihood~\cite{bell1995information} is formulated using maximum likelihood:
\begin{equation}
\log _{\mathcal{X}}(x)=\log _{\mathcal{Z}}(z)+\sum_{k=1}^K \log \left|\operatorname{det}\left(\frac{\partial z_k}{\partial z_{k-1}}\right)\right|,
\end{equation}

\noindent \textbf{ExpFlow.}
\label{sec:3.2.1}
We apply normalizing flow~\cite{Rezende_Mohamed_2015} to produce the expression coefficients of talking face with diverse facial emotion dynamics. However, several non-trivial challenges have emerged: 1) Calculating $\operatorname{det}\left(\frac{\partial z_k}{\partial z_{k-1}}\right)$ comes with computational complexity that approaches $\mathcal{O}(D^3)$, which is intractable for large input dimension $D$. 2) Encoding various emotional coefficients into the latent space and further sampling the specified latent code $z$ has been minimally explored. 3) The scarcity of emotional audio-visual dataset poses difficulties for modeling mixture distributions. To tackle these challenges, we introduce ExpFlow, which not only establishes a more efficient architecture but also leverages a mixture model designed for few-shot learning.

\cref{fig:flow} illustrates the framework of ExpFlow built on~\cite{kingma2018glow, henter2020moglow, ji2023stylevr}. The ground truth (GT) emotional coefficients $\beta^t_e \in \mathbb{R}^{64}$ (for simplicity, we omit time $t$ in the following) and conditional context $c$ pass through $K$ flow steps to yield latent code $z \in \mathbb{R}^{64}$. In our work, the context $c$ contains the source coefficient $\beta_0$ for preserving identity, driving audio $a$ for lip motion guidance, the previous $\tau$ coefficients $\beta^{pre}_e = \beta^{t-\tau:t-1}_e$ for video frame consistency and emotion label $e$ for specifying desired expression.

Each flow step of transformation $f^{-1}_i(\cdot)$ consists of multi-actnorm, invertible convolution and the affine coupling layer. A detailed schematic is provided in the supplementary (\textit{Suppl}). For the multi-actnorm, given the mean $\mu$ and standard deviation $\delta$ for each set of emotional data, we implement it as an affine transformation: $h' = \frac{\beta-\mu}{\delta}$. In our case, we initialize $\mu$ and $\delta$ with the same parameters for each emotion and update them during training, which helps mitigate overfitting~\cite{ji2023stylevr}. Next, ExpFlow introduces an invertible $1 \times 1$ convolution layer: $h'' = \mathbf{W} \cdot h'$, designed to handle potential channel variations. Following this, we utilize a coupling layer based on a transformer $\mathcal{F}$ to generate $h$ from $h''$ and $c$. More specifically, we split $h''$ into $h''_{h1}$ and $h''_{h2}$, where $h''_{h2}$ is affinely transformed by $\mathcal{F}$ based on $h''_{h1}$:
\begin{equation}
    t,s = \mathcal{F}(h''_{h1},c);\qquad  h = [h''_{h1},(h''_{h2}+t)\odot s],
\end{equation}
where $t$ and $s$ denote the transformation parameters. Thanks to the preserved $h''_{h1}$, we can maintain tractability in the reverse direction. To sum up, we have the capability to map $\beta_e$ into the latent code $z$ and predict coefficients based on a sampled code $\hat{z} \in p\mathcal{Z}$ as follows:
\begin{equation}
\label{eq:reverse}
    z = f^{-1}(\beta_e,c); \qquad  \hat{\beta}_e = f(\hat{z},c)
\end{equation}
Furthermore, we mitigate the computational complexity from $\mathcal{O}(D^3)$ by streamlining the computation of the Jacobian determinant using a transformer $\mathcal{F}$, analyzed in~\cite{kingma2018glow}.

Up to this point, the need for a suitable distribution model $p_\mathcal{Z}$ is paramount. To this end, we turn to Student's $t$ Mixture Model (SMM) which encompasses a `fat tail' of multivariate $t$-distribution, particularly effective when working with our relatively small datasets~\cite{alexanderson2020robust}. Concretely, if the outliers within the coefficient distribution cannot be adequately explained by the latent distribution, they exert an unbounded influence on the maximum likelihood process. Therefore, to mitigate the impact of these outlier data points, the $t$-distribution becomes our preferred choice. Given emotion label $e \in {1,\cdots, C}$, mean $\mu_i$, $\Sigma_i$ and the degrees of freedom $\nu(>0)$, a multivariate $t$-distribution and marginal distribution of $z$ known as SMM are expressed as:
\begin{equation}
    p_{\mathcal{Z}}(z \mid e=i)=t_\nu\left(z \mid \mu_i, \Sigma_i\right),
\end{equation}
\begin{equation}
p_{\mathcal{Z}}(z)=\sum_{i=1}^{\mathcal{C}} \pi_i t_\nu\left(z \mid \mu_i, \Sigma_i\right),
\end{equation}
where $\pi_i$ is the mixture coefficient. Here we hypothesise that all categories of emotions have the same proportion, setting $\pi_i=\frac{1}{\mathcal{C}}$. Following~\cite{ji2023stylevr}, the emotion-class-conditional likelihoods of $\beta_e$ is:
\begin{equation}
\label{eq:smm}
p_{\beta_e \sim \mathcal{X}}(\beta_e \mid e=i)=t_\nu\left(z \mid \mu_i, \Sigma_i\right) \cdot \prod_{k=1}^K\left|\operatorname{det}\left(\frac{\partial z_k}{\partial z_{k-1}}\right)\right|
\end{equation}  

\begin{figure*}[t]
  \centering
  \includegraphics[width=0.9\linewidth]{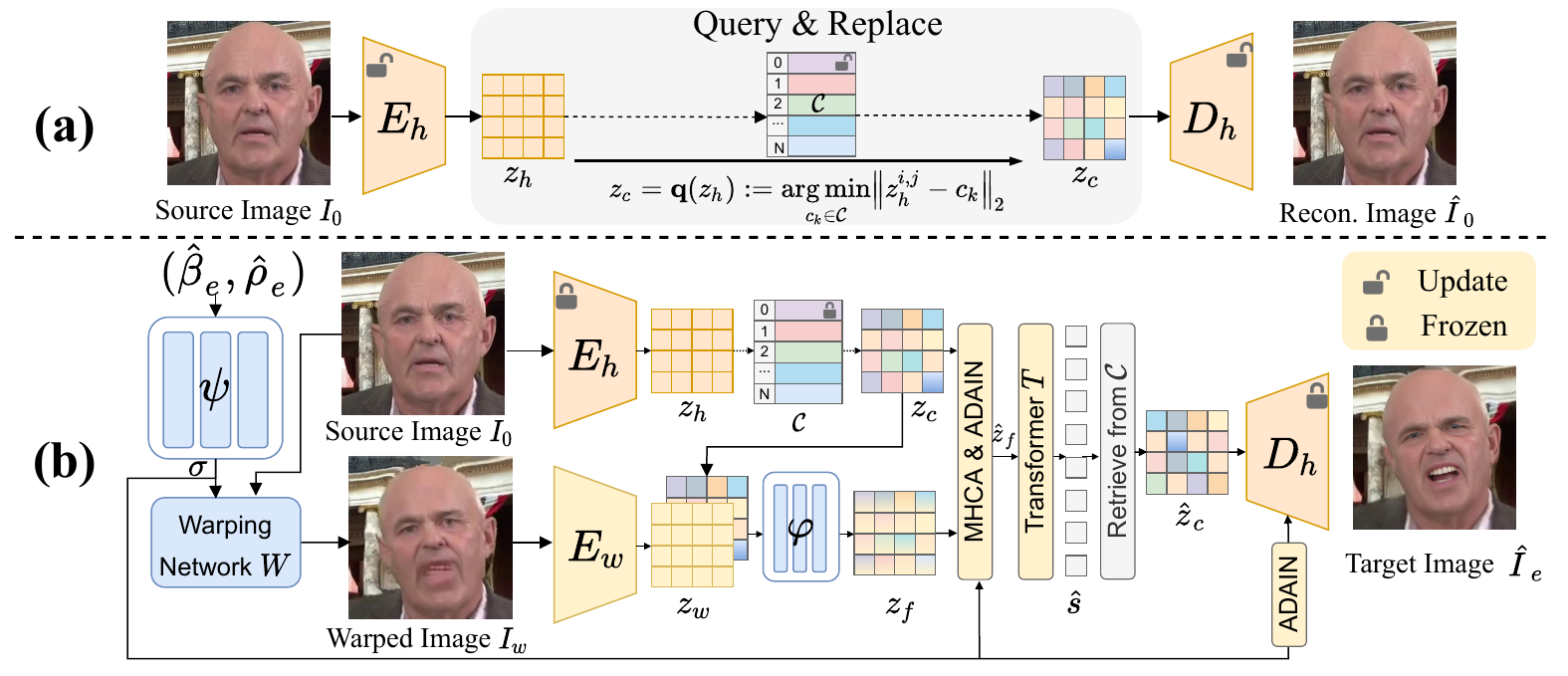}
    \caption{The structure of Vector-Quantized Image Generator (VQIG). (a) Codebook. Initially, we train a high-quality image encoder $E_h$, codebook $\mathcal{C}$ and image decoder $D_h$ using a self-reconstruction strategy. Once converged, we freeze the parameters of ${E_h}$, $\mathcal{C}$ and $D_h$. (b) VQIG. $I_0$ are encoded as a discrete representation $z_c$ to capture identity and texture information. $(\hat{\beta}_e, \hat{\rho}_e)$ generated by~\cref{sec:3.2} are mapped into $\sigma$ by $\psi$, which is then used to warp $I_0$ into $I_w$. An additional warped image encoder $E_w$ is introduced to encode $I_w$ as $z_w$. We employ $\varphi$ to derive $z_f$ from $z_w$ and $z_c$. Along with $z_c$, both are fed into the multi-head cross-attention module (MHCA) and a transformer $T$. The generated vectors retrieve the best-matching features from the Codebook $\mathcal{C}$. The final result is rendered through $D_h$. It is worth noting that we introduce ADAIN at both feature and image levels to incorporate additional spatial information.}
    \label{fig:imagegenerator}
\end{figure*}

To ensure that the mean vectors of different emotions are distinct from each other in the latent space, we randomly initialize the mean vectors $\mu$ of each emotion $i$ from the standard normal distribution $\mu_i \sim \mathcal{N}(0, \textbf{I})$ and maintain their covariance matrices as identity according to~\cite{ji2023stylevr}. In combination with SMM, we can train our ExpFlow by minimizing the negative log-likelihood loss (MLE):
\begin{equation}
\label{eq:3}
\mathcal{L}_\text{exp}=-\sum_{t=0}^{T-1} \log p_\mathcal{Z}\left(f^{-1}(\beta^t_e, c) \mid e^t\right),
\end{equation}
where $T$ is the length of audio $a$. To guarantee video continuity, we introduce the consistency loss $\mathcal{L}_\text{con}$:
\begin{equation}
\label{eq:con}
\mathcal{L}_\text{con}=\|(\beta^{t}_e-\beta^{t-1}_e)-(\hat{\beta}^{t}_e-\hat{\beta}^{t-1}_e)\|_1,
\end{equation}
where $\hat{\beta}_e$ is generated by the reverse process in~\cref{eq:reverse}.

During inference, we produce the coefficient sequence in an autoregressive fashion, with the current output $\hat{\beta}^t_e$ serving as the context for the subsequent iteration, as the previous frame $\beta^{pre}_e$. Furthermore, we have the option to replace the randomly sampled $\hat{z}$ with $\hat{z}^r_e = f^{-1}(\beta^r_e)$ to facilitate emotion transfer, where $\beta^r_e$ are extracted from the emotion reference $I^r_e$, as depicted in~\ref{fig:pipeline}.

We notice that the dropout of $\beta^{pre}_e$ in context $c$ significantly influences emotion diversity (as confirmed in~\cref{fig:latent_space}). Besides, while the 'fat tail' of the $t$-distribution has effectively addressed the outlier issue resulting from limited data, we have further improved performance through Manifold Projection~\cite{lu2021live}. Particularly, we apply the trained ExpFlow to obtain and store all $z \in \mathbb{R}^{64}$ from the training set as the prior $\mathcal{D} \in \mathbb{R}^{N \times 64}$. At inference time, for each randomly sampled $\hat{z}$, we find the $\mathcal{K}$ nearest points $\{\overline{z}_1, \cdots,\overline{z}_\mathcal{K}\}$ in $\mathcal{D}$. Subsequently, we substitute $\sum^\mathcal{K}_{k=1}w_k\cdot \overline{z}_k$ for $\hat{z}$, where the weights $w_k$ are determined by minimizing $\text{min}\|\hat{z} - \sum^\mathcal{K}_{k=1}w_k\cdot \overline{z}_k\|^2_2$, $\sum^\mathcal{K}_{k=1}w_k=1$. In this way, we maintain the original diversity while enhancing robustness.



\noindent \textbf{PoseFlow.}
\label{sec:3.2.2}
PoseFlow is designed to generate a sequence of head poses based on audio and pose history. To achieve this, PoseFlow follows a similar framework as ExpFlow shown in~\cref{fig:flow}. However, there are several key modifications. First, given that the emotional dataset MEAD~\cite{wang2020mead} used in our work, used in our work lacks pose information, we adapt the context $c$ from $[\beta_0, a, \beta^{pre}_e, e]$ to $c_\text{pose} = [a, \rho^{pre}]$. The rationale for excluding the source head pose $\rho_0$ is that a single frame's pose does not convey identity information. Second, we employ a Gaussian distribution $\mathcal{N}(0,\textbf{I})$ as our $p_\mathcal{Z}$ instead of SMM, as we disregard the influence of emotion on head pose. Third, given the pose flow step $f_{pose}$, we replace the loss function in~\cref{eq:3} with $\mathcal{L}_\text{pose}$:
\begin{equation}
    \mathcal{L}_\text{pose}=-\sum_{t=0}^{T-1} \log p_\mathcal{Z}\left(f^{-1}_{pose}(\rho^t, c_{pose})\right)
\end{equation}

\subsection{Vector-Quantized Image Generator}
\label{sec:3.3}
Upon obtaining the emotional coefficients $(\hat{\beta}_e, \hat{\rho}_e)$, we employ them to animate the source image $I_0$ with intricate textures, which are crucial for conveying desired emotions. Our key insight is to build a discrete codebook, which stores high-quality visual textures of face images including teeth, providing essential details to enhance the quality of the animated results. Leveraging this context-rich codebook, we introduce our Vector-Quantized Image Generator (VQIG) to enable high-fidelity and expressive image rendering.

\noindent \textbf{Codebook.} We draw inspiration from VQGAN~\cite{Esser_Rombach_Ommer_2021}, and train our texture-preserving codebook prior by self-reconstruction. Concretely, as illustrated in~\cref{fig:imagegenerator}a, we initially apply a high-quality image encoder $E_h$ to embed the source image $I_0 \in \mathbb{R}^{H \times W \times 3}$ into the latent vector $z_h \in \mathbb{R}^{m\times n\times d}$. Then we generate the vector-quantized representation $z_c$ by involving an introduced codebook $\mathcal{C} = \{c_k\in \mathbb{R}^d\}^N_{k=1}$ and replacing $z_c$ with the queried nearest code $c_k$ in $\mathcal{C}$:
\begin{equation}
    z_c = \textbf{q}(z_h):=\underset{{c_k} \in  \mathcal{C}}{\arg \min }\left\|{z^{i,j}_h}-{c_k}\right\|_2.
\end{equation}
Subsequently, an image decoder $D_h$ is leveraged to reconstruct the input image $\hat{I}_0$. To jointly train the above modules in an end-to-end fashion, we adopt reconstruction loss $\mathcal{L}_\text{rec}$, perceptual loss $\mathcal{L}_\text{per}$~\cite{johnson2016perceptual, zhang2018unreasonable} and adversarial loss $\mathcal{L}_\text{adv}$~\cite{Esser_Rombach_Ommer_2021}:
\begin{equation}
\label{eq:1}
   \mathcal{L}_\text{rec} = \|I_0-\hat{I}_0\|_1; \qquad \mathcal{L}_\text{per} = \|\Phi(I_0)-\Phi(\hat{I}_0)\|^2_2;
\end{equation}
\begin{equation}
\label{eq:2}
\mathcal{L}_\text{adv} = \text{log}D(I_0)+\text{log}(1-D(\hat{I}_0)),
\end{equation}
where $\Phi$ denotes the feature extractor of VGG19~\cite{simonyan2014very}. To update $E_h$ and $\mathcal{C}$, we utilize code-level loss $\mathcal{L}_\text{code}$ and $\mathcal{L}_\text{feat}$:
\begin{equation}
    \mathcal{L}_\text{code} = \|\text{sg}(z_h)-z_c\|^2_2; \qquad \mathcal{L}_\text{feat} = \|z_h-\text{sg}(z_c)\|^2_2,
\end{equation}
where $\text{sg}(\cdot)$ donates the stop-gradient operator. Given the loss weights $\lambda$s, the total loss $\mathcal{L}_\text{tot}$ is represented as:
\begin{equation}
    \mathcal{L}_\text{tot} = \mathcal{L}_\text{rec}+\mathcal{L}_\text{pre}+\lambda_\text{adv}\mathcal{L}_\text{adv}+\mathcal{L}_\text{code}+\lambda_\text{feat}\mathcal{L}_\text{feat}.
\end{equation}

\noindent \textbf{VQIG.} Once the codebook $\mathcal{C}$ is well-trained, we freeze the parameters of $E_h$, $\mathcal{C}$ and $D_h$, and devise Vector-Quantized Image Generator (VQIG) for image animation shown in~\cref{fig:imagegenerator}b. Specifically, the source image $I_0$ is first transformed into $z_h = E_h(I_0)$ and quantize it as $z_c = \textbf{q}(z_h)$, which delivers the identity and texture information. Considering that the predicted coefficients $(\hat{\beta}_e, \hat{\rho}_e)$ provide the motion guidance, we introduce a mapping network $\Phi$~\cite{ren2021pirenderer} to generate motion descriptors $\sigma = \Phi(\hat{\beta}_e, \hat{\rho}_e)$, which are used to further warp $I_0$ as $I_w$ via a warping network $W$. Subsequently, we extract $z_w$ using warped image encoder $E_w$, which is fine-tuned from $E_h$ during training VQIG. While $I_w$ roughly achieves spatial transformation, it may struggle to preserve identity and texture information with detrimental artifacts. To this end, we compensate with $z_c$ by combining it with $z_w$ as $z_f$ using a fuse network $\varphi$. To enhance the performance of emotion-aware texture, we employ a multi-head cross-attention mechanism (MHCA)~\cite{wang2022restoreformer} and an adaptive instance normalization (AdaIN)~\cite{huang2017arbitrary} operator to spatially fuse $z_c$ and $z_f$, which helps to restore the face with fidelity and motion guidance, respectively. Subsequently, we insert a transformer $T$~\cite{zhou2022towards} to predict code sequence $\hat{s} \in \{0,\cdot \cdot \cdot,N-1\}^{m\cdot n}$, which retrieves the respective code items from learned codebook $\mathcal{C}$, forming quantized features $\hat{z}_c$. Through the decoder $D_h$, we generate high-fidelity and expressive face images $\hat{I}_e$ with clear teeth.

To train our VQIG, we adopt the two-stage train strategy~\cite{zhou2022towards}: code-level and image-level supervision. Firstly, we extract code sequence $s$ and latent features $z_c$ from GT frame and minimize the difference with the predicted ones:
\begin{equation}
   \mathcal{L}^\text{VQIG}_\text{code} = \sum^{m\cdot n -1}_{i=0}-s_i\text{log}(\hat{s}_i); \qquad \mathcal{L}^\text{VQIG}_\text{feat} = \|\hat{z}_f-z_c\|^2_2.
\end{equation}

Besides, we incorporate the same image-level loss functions as~\cref{eq:1} and~\cref{eq:2}.
\section{Experiments}
\label{sec:experiments}

\begin{figure*}[t]
  \centering
  \includegraphics[width=0.9\linewidth]{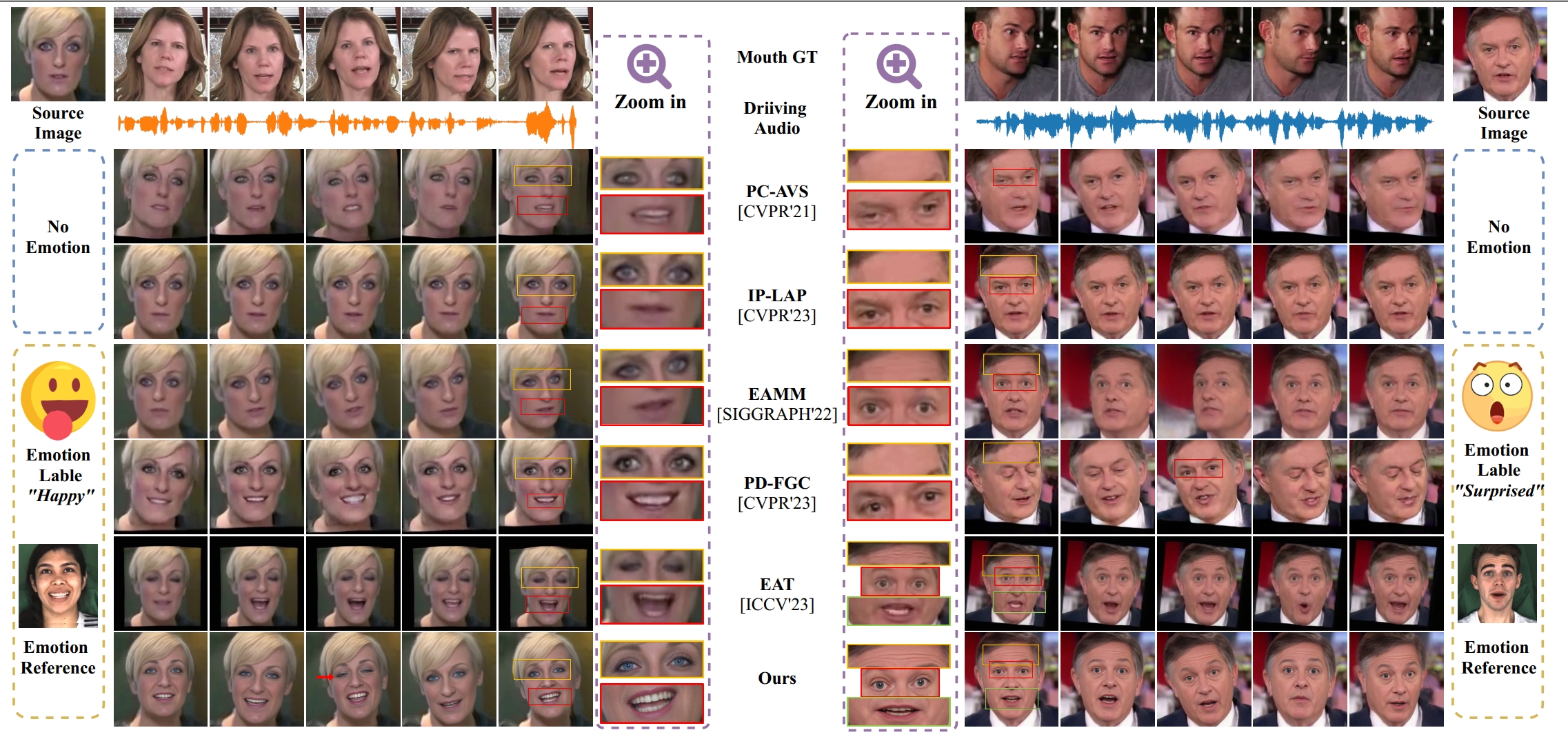}
    \caption{Qualitative comparisons with state-of-the-art methods. See full comparison in supplementary material (\textit{Suppl}).}
    \label{fig:compare1}
\end{figure*}

\subsection{Experimental Settings}
\noindent \textbf{Datasets and Implementation Details.} We train our framework on MEAD~\cite{wang2020mead} and HDTF~\cite{zhang2021flow}. During training codebook, we additionally incorporate FFHQ dataset~\cite{karras2019style} to further improve face modeling capabilities and enhance the preservation of expressive textures. MEAD includes videos of 60 participants expressing 8 different emotions while speaking 30 sentences. HDTF collects various talking videos featuring over 300 identities from YouTube. We categorize the video clips in MEAD into eight emotion categories as originally specified and designate the video clips in HDTF as the ninth category, as they generally represent speaking styles closer to reality without pronounced emotions. FFHQ dataset comprises 70,000 high-quality face images. We crop and resize all data as a resolution of $512 \times 512$ and the latent vector dims are $m = n=16,d=256$. The codebook size and loss weights are set as $N=1024$, $\lambda_\text{adv}=0.8$, $\lambda_\text{feat}=0.25$, respectively. Our method is implemented with PyTorch and trained using the Adam optimizer~\cite{kingma2014adam} on 4 NVIDIA GeForce GTX 3090.


\noindent \textbf{Comparison Setting.}
We compare our method with: (a) emotion-agnostic talking face generation methods: Wav2Lip~\cite{prajwal2020lip}, PC-AVS~\cite{zhou2021pose}, IP-LAP~\cite{zhong2023identity}; (b) emotional talking face generation methods: EAMM~\cite{ji2022eamm}, EMMN~\cite{tan2023emmn},
EAT~\cite{gan2023efficient}, PD-FGC~\cite{wang2023progressive}. The former focuses on the synchronization between the generated lip motion and the input audio, which is validated by Landmarks Distances on the Mouth (M-LMD)~\cite{chen2019hierarchical} and the confidence score of SyncNet~\cite{chung2017out}. In contrast, the emotional talking head generation methods performs expressive expressions on the whole face, where we employ Facial Landmarks Distances (F-LMD) for evaluation. In addition, we fine-tune the Emotion-Fan~\cite{meng2019frame} using MEAD dataset and measure the emotion accuracy ($\text{Acc}_\text{emo}$). Furthermore, SSIM~\cite{ZhouWang2004ImageQA} and FID~\cite{Seitzer2020FID} are adopted to assess image quality, while cumulative probability blur detection (CPBD)~\cite{narvekar2011no} is introduced to evaluate the clarity of the texture.


\begin{table*}[t]
	\centering
	\resizebox{\linewidth}{!}{
	\begin{tabular}{l|cccccc|ccccc|cc|cc}
		\toprule
		\multicolumn{1}{c}{\multirow{2}[4]{*}{\textbf{Method}}} & \multicolumn{6}{c}{\textbf{MEAD}~\cite{wang2020mead}} & \multicolumn{5}{c}{\textbf{HDTF}~\cite{zhang2021flow}}&
		\multicolumn{2}{c}{\textbf{Emotion Input}}&
		\multicolumn{2}{c}{\textbf{Output}}\\
		\cmidrule(lr){2-7}  \cmidrule(lr){8-12} \cmidrule(lr){13-14} \cmidrule(lr){15-16} \multicolumn{1}{c}{} &
		\multicolumn{1}{c}{SSIM$\uparrow$} & \multicolumn{1}{c}{FID$\downarrow$} & \multicolumn{1}{c}{M/ F-LMD$\downarrow$} & \multicolumn{1}{c}{$\text{Sync}_\text{conf}\uparrow$} & \multicolumn{1}{c}{CPBD$\uparrow$} &
		\multicolumn{1}{c}{$\text{Acc}_\text{emo}\uparrow$}
		 &\multicolumn{1}{c}{SSIM$\uparrow$} & \multicolumn{1}{c}{FID$\downarrow$} & \multicolumn{1}{c}{M/ F-LMD$\downarrow$} & \multicolumn{1}{c}{$\text{Sync}_\text{conf}\uparrow$} & \multicolumn{1}{c}{CPBD$\uparrow$} &
		 \multicolumn{1}{c}{Label} &\multicolumn{1}{c}{Reference} &\multicolumn{1}{c}{Diversity} &\multicolumn{1}{c}{HD}
		\\
		\midrule
		\multirow{1}[2]{*}{Wav2Lip~\cite{prajwal2020lip}} & 0.648  & 28.924   &2.294 / 2.234  &  \textbf{8.484}    &   0.104 &17.64\%    &   \textbf{0.742}  &19.757      & 1.767 / 1.732&\textbf{9.073}& 0.126& \XSolidBrush & \XSolidBrush& \XSolidBrush & \XSolidBrush   \\
		\multirow{1}[2]{*}{PC-AVS~\cite{zhou2021pose}} & 0.510 & 36.804    &3.130 / 4.062     &   5.641   &  0.125    &   15.84\%   & 0.690   &   17.617   &1.637 / 2.217    &8.520&0.119& \XSolidBrush & \XSolidBrush& \XSolidBrush & \XSolidBrush   \\

		\multirow{1}[2]{*}{IP-LAP~\cite{zhong2023identity}} & 0.641 & 26.823    &2.303 / 2.205     &  3.371   &  0.116    &   17.61\%   &0.710    &   18.461  & 1.789 / \textbf{1.708}      &3.357&0.142& \XSolidBrush & \XSolidBrush& \XSolidBrush & \XSolidBrush  \\
		\midrule
		\multirow{1}[2]{*}{EAMM~\cite{ji2022eamm}} & 0.621& 26.478    &2.624 / 2.762  &  1.594    & 0.106   & 43.68\%    &  0.604   &27.302    & 2.747 / 2.746 &4.296 &0.118& \XSolidBrush & \Checkmark& \XSolidBrush & \XSolidBrush\\
		\multirow{1}[2]{*}{PD-FGC~\cite{wang2023progressive}} & 0.684 & 27.511 & 2.104 / 2.112 & 5.196 & 0.103 & 61.47\% & 0.692 & 16.929 & 1.720 / 1.966 & 7.321 & 0.128& \XSolidBrush & \Checkmark& \XSolidBrush & \XSolidBrush  \\

		\multirow{1}[2]{*}{EMMN~\cite{tan2023emmn}} &0.675 & 22.895 & 2.439 / 2.851 & 5.125 & 0.116 & 58.53\% & 0.671 & 20.137 & 2.513 / 2.924 & 5.844 & 0.114 & \Checkmark & \XSolidBrush& \XSolidBrush & \XSolidBrush  \\
		\multirow{1}[2]{*}{EAT~\cite{gan2023efficient}} &0.684 & 19.836 & 2.056 / 2.284 & 6.533 & 0.120 & 65.83\% & 0.706 & 18.316 & 1.945 / 2.026 & 7.428 & 0.121& \Checkmark & \XSolidBrush& \XSolidBrush & \XSolidBrush   \\ 
    	\midrule
		\textbf{FlowVQTalker} & \textbf{0.689} & \textbf{16.553} & \textbf{1.939} / \textbf{2.061} & 5.901 & \textbf{0.181} & \textbf{71.53\%} & 0.708 & \textbf{15.165} & \textbf{1.643} / 1.958 & 6.766 & \textbf{0.268} & \Checkmark& \Checkmark& \Checkmark& \Checkmark   \\ 
		\midrule
		\multirow{1}[2]{*}{GT} & 1.000 & 0.000 & 0.000 / 0.000 & 6.733 & 0.161 & 81.68\% & 1.000 & 0.000 & 0.000 / 0.000 & 7.728 & 0.238&-  &-&-&-\\
		\bottomrule
	
	\end{tabular}%
	}
	\caption{Quantitative comparisons with state-of-the-art methods. Supplementary material gives more quantitative comparison results.}
	\label{tab:quantitative}%
\end{table*}%

\subsection{Compare with other state-of-the-art methods}
\noindent \textbf{Talking Face Generation.}
~\cref{fig:compare1} displays video frames generated by various SOTA methods (See \textit{Suppl} for full comparison). PC-AVS and IP-LAP struggle with preserving identity information and lip synchronization, respectively. Additionally, both methods cannot generate videos with emotional expressions. Although EAMM resorts to a driving video as emotion guidance, it fails to perform vivid expression on the whole face. PD-FGC and EAT can predict happy faces, but PD-FGC's surprised faces lack clear expressions, and EAT faces issues with closed eyes. Our results demonstrate the preservation of identity information while accurately expressing corresponding emotions. Note that since $\hat{z}$ is randomly sampled from the modeled distribution, our method can randomly generate blinks (as pointed out by red arrow), contributing to expressive and dynamic talking faces. Furthermore, please see the zoom-in details, the compared methods struggle to synthesize fine-grained, emotion-aware textures and clear teeth. In contrast, our method excels at generating high-quality images, even when the source image is blurred.~\cref{tab:quantitative} shows that our FlowVQTalker achieves the best performance across most evaluation criteria. Wav2Lip~\cite{prajwal2020lip} obtains the highest scores in $\text{Sync}_\text{conf}$ which even surpasses the ground truth (GT). We assume that Wav2Lip uses SyncNet confidence as a critical constraint using a SyncNet discriminator~\cite{chung2017out}, which is reasonable to pursue a higher $\text{Sync}_\text{conf}$. We obtain similar scores to GT and lower M-LMD, demonstrating our ability to predict synchronized lip motions. Moreover, since Wav2Lip and IP-LAP only edit mouth region and keep other facial parts unchanged, they achieve highest SSIM and lowest F-LMD, respectively. Thanks to our texture-rich codebook, FlowVQTalker significantly outperforms SOTAs regarding CPBD on both datasets, which suggests that the details in our results are clearer and more visually appealing. Note that our emotion source can be emotion label for high-definition (HD) diverse facial emotion dynamics generation, or be an emotion reference for emotion transfer. In contrast, the compared methods can only accommodate one of these inputs, resulting in a deterministic output.

\begin{figure}[t]
  \centering
  \begin{subfigure}{0.8\linewidth}
    \includegraphics[width=1\linewidth]{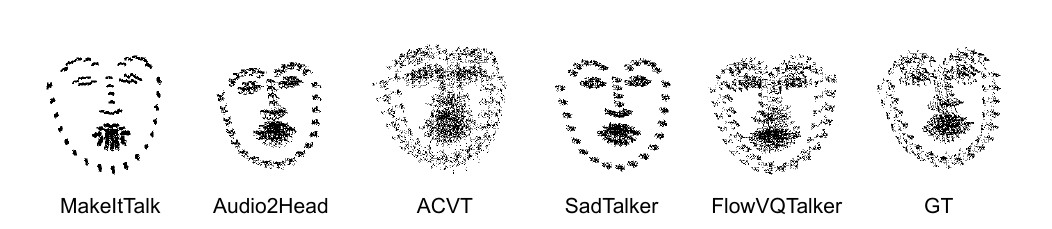}
    \caption{Trace maps}
    \label{fig:pose2_1}
  \end{subfigure}
  \hfill

  \begin{subfigure}{0.9\linewidth}
\includegraphics[width=1\linewidth]{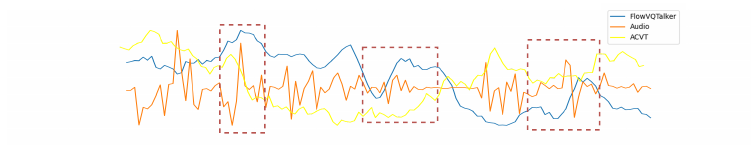}
 \caption{Correlation between audio and poses.}
    \label{fig:pose2_2}
  \end{subfigure}
  \caption{Comparison with SOTAs in terms of generated poses.}
  \label{fig:pose2}
  
\end{figure}

\noindent \textbf{Diversity.} We present facial dynamics diversity and emotion transfer in supplementary video, encompassing expression, blink and pose. Here, we compare with MakeItTalk~\cite{zhou2020makelttalk}, Audio2Head~\cite{wang2021audio2head}, AVCT~\cite{wang2022one} and SadTalker~\cite{zhang2022sadtalker} concerning pose diversity and correlation with audio, assessed using trace map~\cite{zhang2022sadtalker} and correlation map~\cite{wang2021audio2head}.~\cref{fig:pose2_1} demonstrates that only AVCT performs comparable diversity with our FlowVQTalker, while ours achieves better synchronization than ACVT in~\cref{fig:pose2_2}.

\begin{table}[t]
  \centering
  \resizebox{\linewidth}{!}{

  \begin{tabular}{@{}l|cccccc@{}}
    \toprule
    Metric/Method & PC-AVS  & IP-LAP  & EAMM& EAT &FlowVQTalker &GT \\
    \midrule
    Lip-sync & 3.89 &3.91  &3.43&3.94&\textbf{4.03}&4.88 \\
    Iamge-quality & 3.24 & 3.83  &3.46 &3.72&\textbf{4.25}& 4.67 \\
    $\text{Acc}_\text{emo}$ & 31.5\% & 54.9\%  &53.2\% &52.4\%&\textbf{60.6\%}& 76.4\% \\
    \bottomrule
  \end{tabular}
  }
  \caption{User study results.}
  \label{tab:user_study}
\end{table}

\noindent \textbf{User Study.}
We conduct user study to evaluate our method from a human perspective. We recruit 20 participants (10 males/10 females) to score 120 videos (20 videos $\times$ (5 methods + GT)) from 1 (worst) to 5 (best) in terms of lip synchronization and image quality. They are also required to classify the emotion performed by videos. The results, detailed in~\cref{tab:user_study}, clearly illustrate the superiority of our method across all evaluated aspects with the aid of our human-like observations and corresponding solutions.

\begin{figure}[t]
  \centering
  \begin{subfigure}{0.32\linewidth}
\includegraphics[width=1\linewidth]{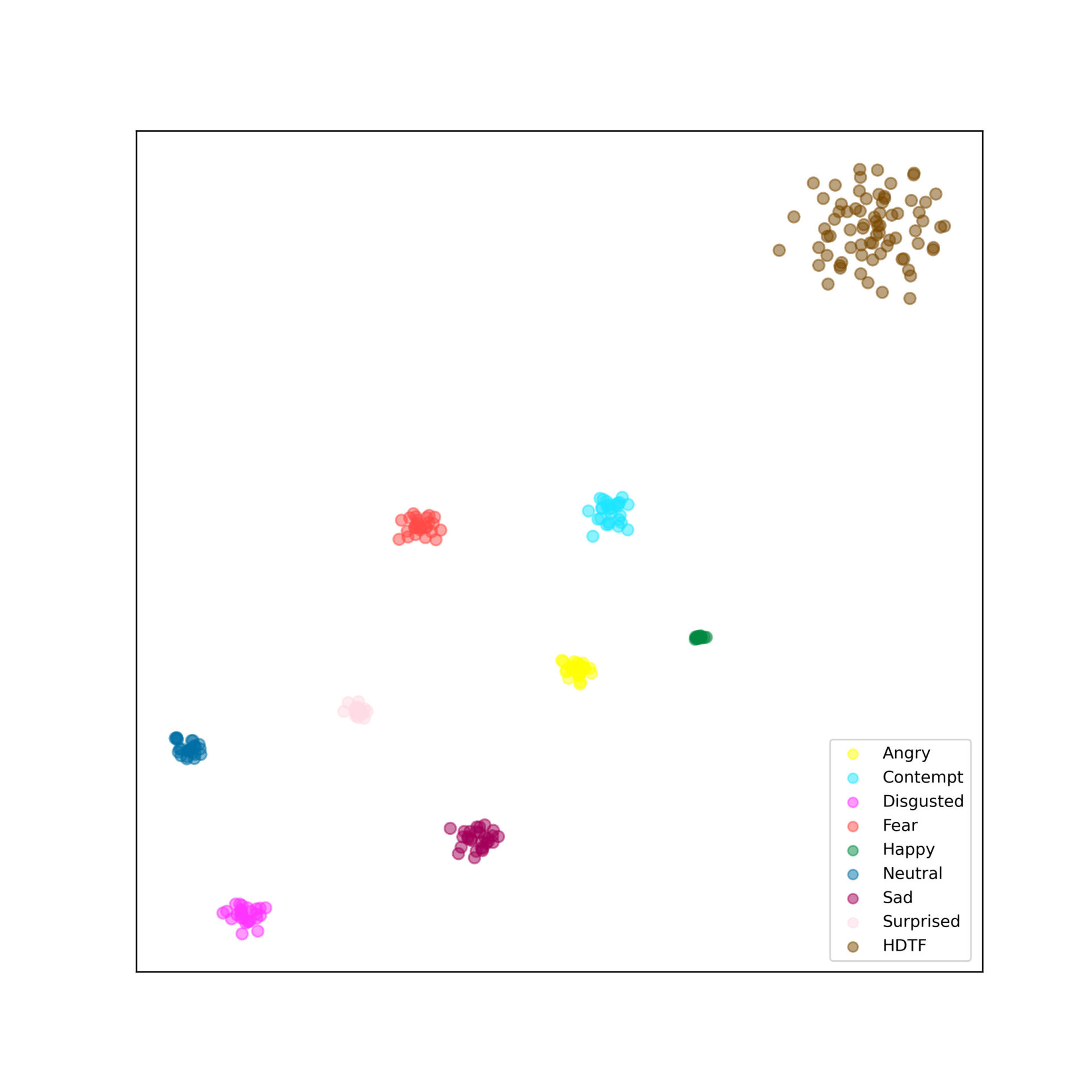}
 \caption{GMM}
    \label{fig:gmm}
  \end{subfigure}
  \hfill
  \begin{subfigure}{0.32\linewidth}
    \includegraphics[width=1\linewidth]{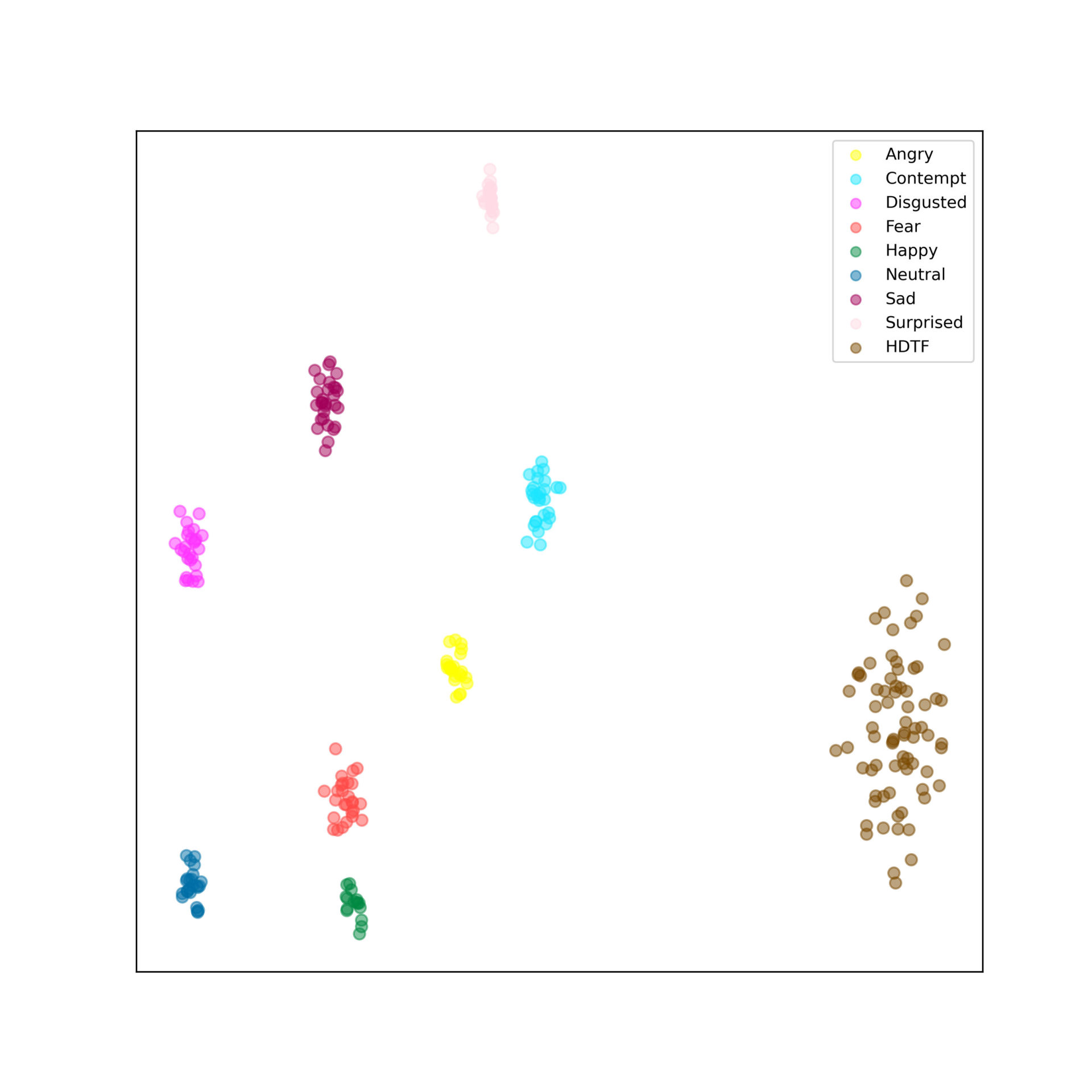}
    \caption{w/o data dropout}
    \label{fig:size1}
  \end{subfigure}
  \hfill
  \begin{subfigure}{0.32\linewidth}
\includegraphics[width=1\linewidth]{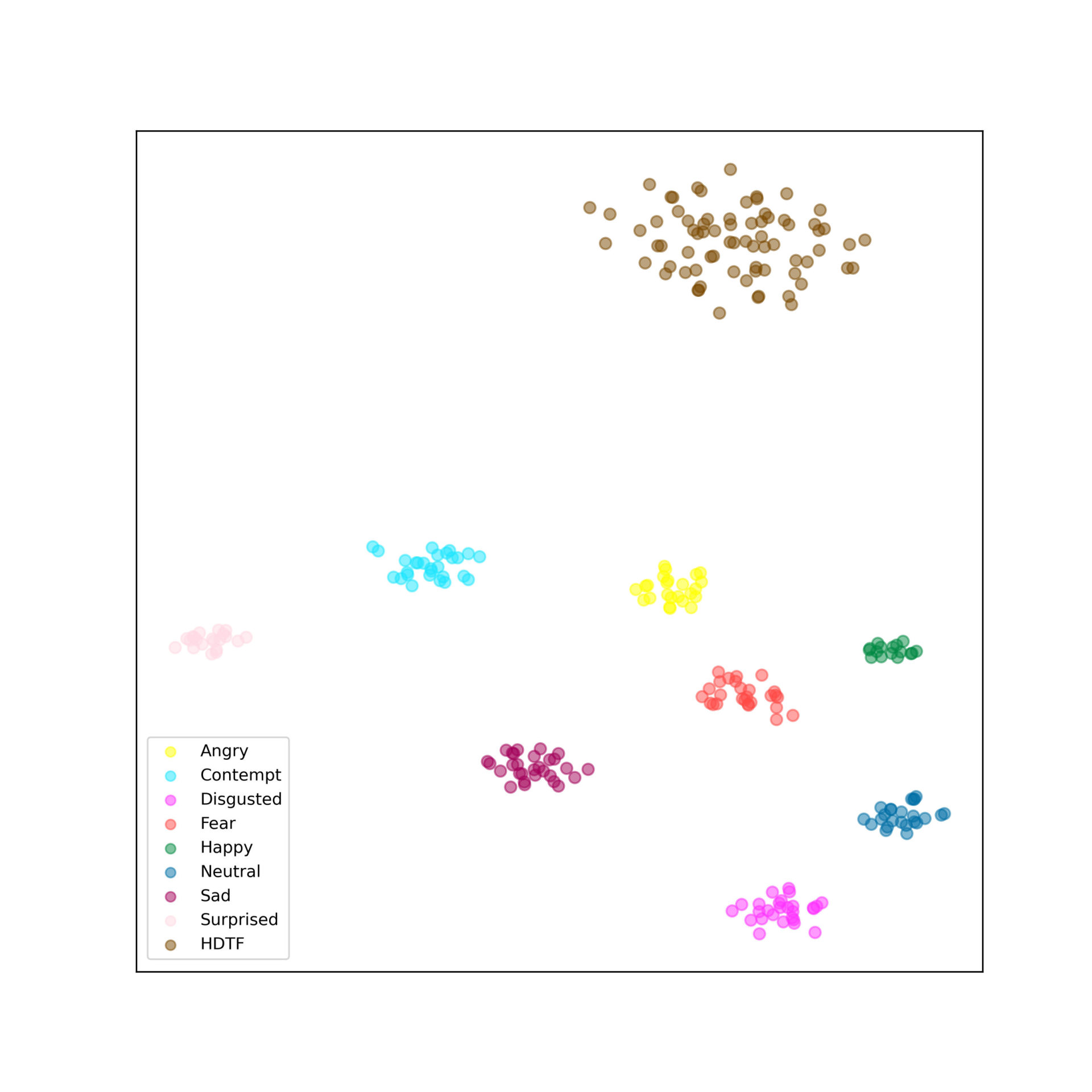}
 \caption{Ours}
    \label{fig:ours}
  \end{subfigure}
  \caption{Visualization of latent space.}
  \label{fig:latent_space}
\end{figure}

\subsection{Ablation Study.}
\noindent \textbf{Ablation of ExpFlow.} For ExpFlow, we delve into the impact of different settings on the latent space, a crucial element in emotional modeling and diversity. We examin two key variations: (1) \textbf{GMM}: replace SMM with Gaussian Mixture Model (GMM). (2) \textbf{w/o data dropout}: use SMM but exclude data dropout.~\cref{fig:latent_space} presents the visualization of the latent space. GMM tends to overfit and form clusters around specific data points within our relatively small dataset~\cite{wang2020mead}. This leads to poor performance as the model is prone to sampling outliers. While SMM, with its long-tailed distributions, mitigates this 
issue, data dropout is also critical for constructing more resilient distributions. In addition, data dropout also improves the consistency between the generated motion and the context, achieving better synchronization of audio and lip motion as shown in \textit{Suppl}.
\begin{figure}[t]
  \centering
  \includegraphics[width=0.85\linewidth]{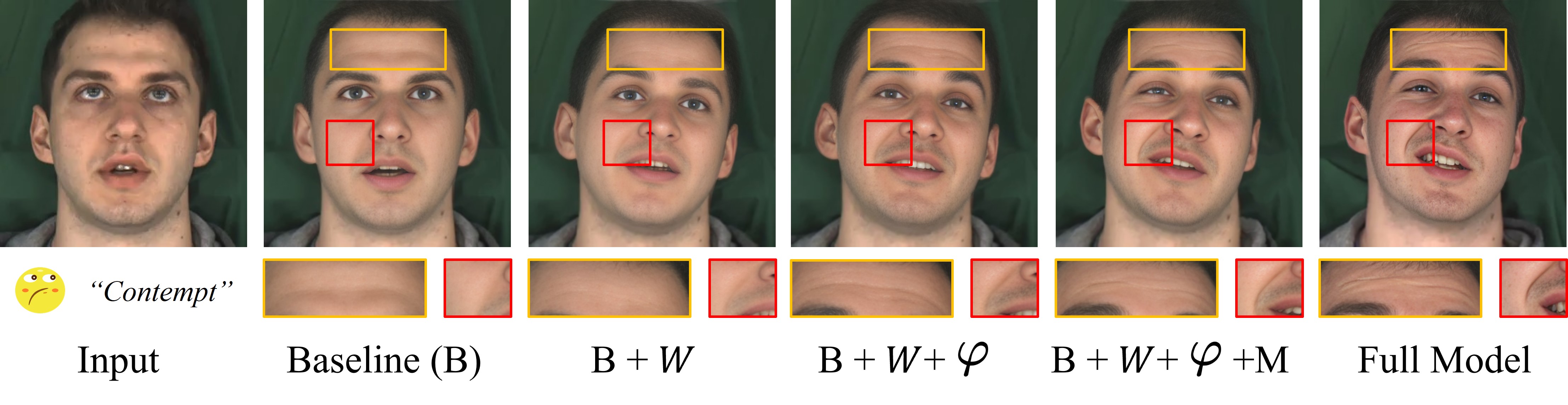}
    \caption{Visualization results of ablation study.}
    \label{fig:ablation}
\end{figure}

\noindent \textbf{Ablation of VQIG.}
We conduct an ablation study on VQIG with following variants: We started with a \textbf{baseline (B)}, which retains only $E_w$, ADAIN and $D_h$. Subsequently, we add $W$, $\varphi$, MHCA and $\mathcal{C}$ in turn to verify the validity of each module, namely \textbf{B+$W$}, \textbf{B+$W$+$\varphi$}, \textbf{B+$W$+$\varphi$+M} and \textbf{B+$W$+$\varphi$+M+$\mathcal{C}$} (\textbf{Full Model}). The results presented in~\cref{fig:ablation}, where $W$ initially warps the source image but lacks detail information, $\varphi$ and MHCA effectively fuse identity, texture and spatial information, and the inclusion of $\mathcal{C}$ further improves the fidelity and clarity of the image.

\section{Conclusion}
In this paper, we introduce FlowVQTalker, a system capable of generating talking face with high-definition expression and non-deterministic facial dynamics, addressing both insights we set out. Within FlowVQTalker, flow-based coeff. generator establishes an invertible mapping between coefficients of 3DMM and a distribution model, allowing for random sampling to ensure diverse nonverbal facial expressions. Vector-quantized image generator resorts to a texture-rich codebook and synthesizes realistic videos with fine-grained details, such as emotion-aware wrinkles and clear teeth. We conduct comprehensive experiments to illustrate the superiority of our FlowVQTalker.

\section{Acknowledgments}

This work was supported by National Natural Science Foundation of China
(NSFC, NO. 62102255), Shanghai Municipal Science and
Technology Major Project (No. 2021SHZDZX0102). Tencent Industry-University Collaboration Research Funding (TencentJR2024IEG111). We would like to thank Xinya Ji, Yifeng Ma and Yuan Gan for their generous help.

{
    \small
    \bibliographystyle{ieeenat_fullname}
    \bibliography{main}
}

\end{document}